\documentclass{ecai}
\usepackage{graphicx}
\usepackage{latexsym}

\usepackage{xspace}
\usepackage{xcolor}
\usepackage{amsmath}
\usepackage{amssymb}
\usepackage{todonotes}
\usepackage{paralist}
\usepackage{subcaption}

\newcommand{\myi}{(\emph{i})\xspace}
\newcommand{\myii}{(\emph{ii})\xspace}
\newcommand{\myiii}{(\emph{iii})\xspace}

\newcommand{\mya}{(\emph{a})\xspace}
\newcommand{\myb}{(\emph{b})\xspace}

\newcommand{\tiff}{\text{ iff }}
\newcommand{\LTLftitle}{\textbf{LTL$_f$}\xspace}
\newcommand{\LTLf}{{\sc ltl}$_f$\xspace}

\newcommand{\LTL}{{\sc ltl}\xspace}
\newcommand{\trace}{\pi} 
\newcommand{\ptrace}{\tau} 
\newcommand{\pmodels}{\models_{\D}}

\newcommand{\last}{\mathtt{lst}}

\newcommand{\true}{\mathit{true}}

\newcommand{\false}{\mathit{false}}

\newcommand{\Next}{\raisebox{-0.27ex}{\LARGE$\circ$}}
\newcommand{\Wnext}{\raisebox{-0.27ex}{\LARGE$\bullet$}}
\newcommand{\Until}{\mathop{\U}}

\newcommand{\A}{\mathcal{A}}
\newcommand{\E}{\mathcal{E}}

\renewcommand{\P}{\mathcal{P}} 
\newcommand{\D}{\mathcal{D}}
\newcommand{\U}{\mathcal{U}}

\newcommand{\F}{\mathcal{F}}
\newcommand{\T}{\mathcal{T}}
\renewcommand{\L}{\mathcal{L}}
\renewcommand{\H}{\mathcal{H}}

\newcommand{\DFA}{{DFA}\xspace}
\newcommand{\DFAs}{{DFA}s\xspace}
\newcommand{\acc}{Acc\xspace}

\newcommand{\total}{\rightarrow}
\renewcommand{\part}{\mapsto}

\renewcommand{\implies}{\supset}

\newcommand{\rseq}{\vec{r}}
\newcommand{\aseq}{\vec{a}}
\newcommand{\Trace}{\mathtt{Trace}}
\newcommand{\Play}{\mathtt{Play}}
\newcommand\run[1]{{\sf{Run}{#1}}}

\newcommand{\act}{Act}
\newcommand{\react}{React}

\newcommand{\val}{val}

\newcommand{\agerr}{s^{ag}_{err}}
\newcommand{\enverr}{s^{env}_{err}}
\newcommand{\bigagerr}{S^{ag}_{err}}
\newcommand{\bigenverr}{S^{env}_{err}}
\newcommand{\reach}{\mathtt{Reach}}
\newcommand{\safe}{\mathtt{Safe}}
\newcommand{\safereach}{\mathtt{SafeReach}}

\newcommand{\fond}{FOND\xspace}

\newcommand{\lydia}{{\sc lydia}\xspace}
\newcommand{\besyftp}{\textit{BeSyftP}\xspace}

\newtheorem{definition}{Definition}
\newtheorem{lemma}{Lemma}
\newtheorem{proposition}{Proposition}

\makeatletter
\DeclareRobustCommand{\qed}{%
  \ifmmode 
  \else \leavevmode\unskip\penalty9999 \hbox{}\nobreak\hfill
  \fi
  \quad\hbox{\qedsymbol}}
\newcommand{\openbox}{\leavevmode
  \hbox to.77778em{%
  \hfil\vrule
  \vbox to.675em{\hrule width.6em\vfil\hrule}%
  \vrule\hfil}}
\newcommand{\qedsymbol}{\openbox}

\newcommand{\proofname}{Proof}
\makeatother

\begin{document}

\begin{frontmatter}

\title{\LTLf Best-Effort Synthesis in Nondeterministic \\Planning Domains}



\author[A,B]{\fnms{Giuseppe}~\snm{De Giacomo}}
\author[B]{\fnms{Gianmarco}~\snm{Parretti}\thanks{Corresponding Author. Email: parretti@diag.uniroma1.it}}
\author[A]{\fnms{Shufang}~\snm{Zhu}\thanks{Corresponding Author. Email: shufang.zhu@cs.ox.ac.uk}} 

\address[A]{University of Oxford, UK}
\address[B]{University of Rome ``La Sapienza”, Italy}

\begin{abstract} 
We study best-effort strategies (aka plans) in fully observable nondeterministic domains (\fond) for goals expressed in Linear Temporal Logic on Finite Traces (\LTLf). The notion of best-effort strategy has been introduced to also deal with the scenario when no agent strategy exists that fulfills the goal against every possible nondeterministic environment reaction. Such strategies fulfill the goal if possible, and do their best to do so otherwise. We present a game-theoretic technique for synthesizing best-effort strategies that exploit the specificity of nondeterministic planning domains. We formally show its correctness and demonstrate its effectiveness experimentally, exhibiting a much greater scalability with respect to a direct best-effort synthesis approach based on re-expressing the planning domain as generic environment specifications.
\end{abstract}

\end{frontmatter}

\section{Introduction}\label{sec:intro}
Recently there has been quite some interest in synthesis~\cite{PR89,finkbeiner2016synthesis} for realizing goals (or tasks) $\varphi$ against environment specifications $\E$ \cite{ADMR18,AGMR19}, especially when both $\varphi$ and $\E$ are expressed in Linear Temporal Logic on finite traces~(\LTLf)~\cite{DegVa13,DegVa15}, the finite trace variant of \LTL~\cite{Pnu77}, a logic specification language that is commonly adopted in Formal Methods~\cite{BaKG08}. In this setting, synthesis amounts to finding an agent strategy that wins, i.e., generates a trace satisfying $\varphi$, whatever is the (counter-)strategy chosen by the environment, which in turn has to satisfy its specification $\E$. This form of synthesis can be seen as an extension of FOND planning \cite{GeBo2013,ghallab2004automated}, as shown in, e.g., \cite{DeGR18,CMc19}.

Obviously, a winning strategy for the agent may not exist. To handle this possibility, the notion of strong cyclic plans was introduced~\cite{CimattiRT98,Cimatti03}: if a (strong) plan does not exist, there may still exist a plan that could win assuming the environment is not strictly adversarial.  Building on this intuition, Aminof et al.~\cite{ADLMR20} proposed the notion of best-effort strategies (or plans), which formally capture the idea that the agent could do its best by adopting a strategy that wins against a maximal set (though not all) of possible environment strategies.  It was then shown in \cite{ADR2021} that best-effort strategies capture the game-theoretic rationality principle that a player~(the agent) would not use a strategy that is ``dominated'' by another one~(i.e., if another strategy fulfills the goal against more environment behaviors, then the player should adopt that strategy). Best-effort strategies have some notable properties:  \myi they always exist, \myii if a winning strategy exists, then best-effort strategies are exactly the winning strategies, \myiii best-effort strategies can be computed in 2EXPTIME as winning strategies (best-effort synthesis is indeed 2EXPTIME-complete).

In~\cite{ADR2021} an algorithm for \LTLf best-effort synthesis has been presented. This algorithm is based on creating, solving, and combining the solutions of three distinct games (with three different objectives) played in the same game arena. The arena is obtained from the deterministic finite-state automata (\DFAs) corresponding to the \LTLf specification of the agent goal $\varphi$ and the \LTLf specification of the environment $\E$. As a result, the size of the arena is, in the worst-case, double-exponential both in $\varphi$ and in $\E$. Using this framework, we can also capture best-effort synthesis in nondeterministic planning domains. In particular, one can simply re-express nondeterministic planing domains (FOND) in \LTLf ~\cite{DegVa13,ADMR18,KWKLV19} and then use the \LTLf best-effort synthesis approach directly.
However, observe that in planning, while the (temporally extended) goal $\varphi$ is typically small, the environment specification $\E$ is  large, being the entire planning domain (i.e., a representation of how the world the agent is immersed in works). 
This observation motivates our paper. 


We study \LTLf best-effort synthesis directly in the context of nondeterministic (adversarial) planning domains.
Specifically, our contributions are:
\begin{compactitem}
\item 
A framework for best-effort synthesis in nondeterministic (adversarial) planning domains;
\item 
A synthesis technique for best-effort synthesis, inspired by the one in \cite{ADR2021}, that takes full advantage of the specificity of planning domains as environment specifications;
\item 
A symbolic best-effort synthesis algorithm that performs and scales well;
\item 
An empirical evaluation of the practical effectiveness of the approach.
\end{compactitem}
Our results show that computing best-effort strategies for \LTLf goals in nondeterministic domains is much more effective than using \LTLf best-effort synthesis directly. 
In fact, our technique can be implemented quite efficiently, with  only a small overhead wrt to computing winning strategies (i.e., strong plans) in FOND. 
Hence, it is completely feasible in practice to return a best-effort strategy instead of giving up when a winning strategy does not exist.
\begin{figure}[t]
     \centering
     \includegraphics[scale = .5]{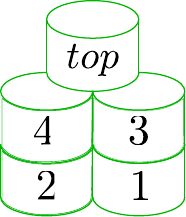}
         \caption{An arch. We denote location names in the blocks.}
    \label{fig:arch}
\end{figure}

\noindent\textbf{Running Example.~\label{sec:1-1}}
We now present a relatively simple example to illustrate the key structure of \LTLf best-effort synthesis in nondeterministic planning domains. This example, adapted from~\cite{KWKLV19}, represents a human-robot co-assembly task in a shared workplace. Specifically, the human and robot involved in the task are considered as the environment and the agent, respectively. 
In the shared workplace, the robot can perform grasp/place actions to pick/drop blocks and transfer/transit actions to move its robotic arm with/without a block in its gripper. After every robot action, the human can react by also moving blocks among locations to interfere with the robot, hence introducing nondeterminism to robot actions. 
%
%
%
Consider a robot \emph{goal} (aka \emph{task}) of assembling an arch of blocks, as depicted in Figure~\ref{fig:arch}. It is easy to see that the agent has \emph{no} \emph{winning strategy} (aka \emph{strong solution}) to assemble the arch, since the human can always disassemble it. Therefore, standard synthesis~\cite{DegVa15,DeGR18} would conclude the task as \emph{unrealizable}, hence ``giving up". 
%
%
However, the robot still has the chance to fulfill the goal, should the human cooperate or even perform flawed reactions due to e.g. lack of adequate training.
Therefore, instead of simply giving up, the agent should try its best to pursue the goal by exploiting human reactions. 
\emph{Best-effort strategies} (aka \emph{best-effort solutions}) precisely capture this intuition.

\section{Preliminaries}\label{sec:2}

\noindent\textbf{Traces.} Let $\Sigma$ be a set of propositions. A \textit{trace} $\trace = \trace_0\trace_1\ldots$ is a sequence of propositional interpretations~(sets), where for every $i \geq 0$, $\trace_i \in 2^{\Sigma}$ is the $i$-th interpretation of $\trace$. Intuitively, $\trace_i$ is interpreted as the set of propositions that are $true$ at instant $i$.  The length of a trace is $|\pi|$. A trace $\trace$ is an \textit{infinite} trace if $|\pi| = \infty$, which is formally denoted as $\trace\in (2^{\Sigma})^{\omega}$; otherwise $\trace$ is a \textit{finite} trace, denoted as $\trace\in (2^{\Sigma})^{*}$. If a trace $\pi$ is finite, we denote by $\last(\pi)$ its last instant~(i.e., index).
Moreover, by $\trace^k = \trace_0 \cdots \trace_k$ we denote the \emph{prefix} of $\trace$ up to the $k$-th iteration.
%



\noindent\textbf{\LTLftitle Basics.} 
\textit{Linear Temporal Logic on finite traces}~(\LTLf) is a specification language to express temporal properties on finite and non-empty traces~\cite{DegVa13}. In particular, \LTLf
has the same syntax as \LTL, which is instead interpreted over infinite traces~\cite{Pnu77}. Given a set of propositions $\Sigma$, \LTLf formulas are generated as follows: \\
{\centerline{$\varphi ::= p \mid \varphi \wedge \varphi \mid \neg \varphi \mid  
 \Next \varphi \mid \varphi \Until \varphi.$}}
$p \in \Sigma$ is an \textit{atom}, $\Next$~(\emph{Next}), and $\Until$~(\emph{Until}) are temporal operators. 
We use standard Boolean abbreviations such as $\vee$~(or) and $\supset$~(implies), $\true$ and $\false$. Moreover, we define the following abbreviations \emph{Weak Next} $\Wnext \varphi \equiv \neg \Next \neg \varphi$, \emph{Eventually} $\Diamond \varphi \equiv \true \Until \varphi$ and \emph{Always} $\Box \varphi \equiv \neg \Diamond \neg \varphi$.
The size of $\varphi$, written $|\varphi|$, is the number of all subformulas of $\varphi$.

Given an \LTLf formula $\varphi$ over $\Sigma$ and a finite, non-empty trace $\trace \in (2^\Sigma)^+$, we define when $\varphi$ \emph{holds} at instant $i~(0 \leq i \leq \last(\pi))$, written as $\trace, i \models \varphi$, inductively on the structure of $\varphi$, as:
\begin{compactitem}
	\item 
	$\trace, i \models p \tiff p \in \trace_i$;
	\item 
	$\trace, i \models \lnot \varphi \tiff \trace, i \not\models \varphi\nonumber$;
	\item 
	$\trace, i \models \varphi_1 \wedge \varphi_2 \tiff \trace, i \models \varphi_1 \text{ and } \trace, i \models \varphi_2$;
	\item 
	$\pi, i \models \Next\varphi \tiff  i< \last(\pi)$ and $\trace,i+1 \models \varphi$;
	\item 
	$\pi, i \models \varphi_1 \Until \varphi_2$ iff $\exists j$ such that $i \leq j \leq \last(\pi)$ and $\pi,j \models\varphi_2$, and $\forall k, i\le k < j$ we have that $\pi, k \models \varphi_1$.
\end{compactitem}

We say $\pi$ \emph{satisfies} $\varphi$, written as $\pi \models \varphi$, if $\pi, 0 \models \varphi$.

\noindent\textbf{FOND Planning for \LTLf Goals.} 
Planning in \textit{Fully Observable Nondeterministic} (FOND) domains for \LTLf goals concerns computing a strategy to fulfill a temporally extended goal expressed as an \LTLf formula in a planning domain, where the agent has full observability, regardless of how the environment non-deterministically reacts to agent actions.
In this paper, we extend existing works on synthesis-based approaches to planning in FOND domains~\cite{DeGR18,CBM19,CMc19} to compute best-effort strategies~\cite{ADR2021}.

\section{Framework~\label{sec:3}}
We begin by presenting our framework. 
First, we introduce the notion of nondeterministic planning domain and then formalize the problem of \LTLf best-effort synthesis in nondeterministic planning domains.

\subsection{Nondeterministic Planning Domains~\label{sec:3-1}}

In our framework, a nondeterministic planning domain is intuitively considered as the arena of a \textit{two-player game}, in which the agent and the environment can perform actions and reactions, respectively. Starting from the initial state of the domain, the agent and the environment move in turns, such that at each turn, the agent makes an action and the environment responds with some reaction. State transitions are determined only if both players complete their moves following the preconditions of their respective moves. In such a domain, the \textit{nondeterminism} for the agent comes from not knowing how the environment will react.

Formally, we define a nondeterministic planning domain as a tuple \(\D = (2^\F, s_0, Act, React, \alpha, \beta, \delta)\), where: $\F$ is a finite set of fluents such that $2^\F$ is the state space and $|\F|$ is the size of the domain; $s_0 \in 2^\F$ is the initial state; $\act$ and $\react$ are finite sets of agent actions and environment reactions, respectively; $\alpha: 2^\F \total 2^{Act}$ and $\beta: 2^\F \times Act \total 2^{React}$ denote preconditions of agent actions and environment reactions, respectively; and $\delta: 2^\F \times Act \times React \part 2^\F$ is the transition function such that \(\delta(s, a, r) = s' \in 2^\F\) if \(a \in \alpha(s)\) and \(r \in \beta(s, a)\), and $\delta(s, a, r)$ is undefined otherwise.
 

In particular, we require planning domains to satisfy the following three rules: \begin{compactitem}
    \item 
    \emph{Existence of agent action}. For every state $s \in 2^\F$, there exists at least one agent action $a$ such that $a \in \alpha(s)$. Formally:
    {\centerline{$\forall s \in 2^\F. \alpha(s) \neq \emptyset.$}}
    This rule guarantees that the agent can perform at least one action in every state of the domain\footnote{If this is not the case, it is sufficient to add a new agent action $\mathit{nop}$ with a corresponding environment reaction $\mathit{nopr}$ such that $\delta(s,\mathit{nop},\mathit{nopr}) = s$.}.
    \item 
    \emph{Existence of environment reaction}. For every state \(s \in 2^\F\) and agent action \(a \in \alpha(s)\), there exists at least one environment reaction \(r \in \beta(s, a)\) such that \(\delta(s, a, r)\) is defined. Formally:
    {\centerline{$\forall s \in 2^\F, \forall a \in \alpha(s). \beta(s, a) \neq \emptyset.$}}
    This rule guarantees that the environment is able to respond to any agent action that follows the action precondition.
    \item 
    \emph{Uniqueness of environment reaction}.
   For every state $s \in 2^\F$, agent action $a \in \alpha(s)$, and successor state $s' = \delta(s, a, r)$ for some $r \in \beta(s, a)$, the reaction $r$ is unique. Formally:
    {\centerline {$\forall s \in 2^\F, \forall a \in \alpha(s), \forall r_1, r_2 \in \beta(s, a).$}} {\centerline{$\delta(s, a, r_1) = \delta(s, a, r_2) \implies r_1 = r_2.$}}
    This rule guarantees that, given a state $s$ and an agent action $a \in \alpha(s)$, all the environment responses in $\beta(s,a)$ are distinguishable by just looking at the resulting successor state\footnote{Observe that if this rule does not hold in our domain, we can easily modify the domain by introducing fluents to represent possible environment reactions and record the reaction in the resulting state, hence complying to the rule.}. 
\end{compactitem}

Observe that the nondeterministic planning domains adopted in FOND~\cite{CimattiRT98,GeBo2013}, say expressed in PDDL~\cite{Haslum2019}, can be immediately captured by our notion along the line discussed in~\cite{DeLe2021}.

A \textit{state trace} of \(\D\) is a (finite or infinite) sequence \(\tau = s_0 s_1 \ldots\) of states in \(2^\F\) such that \(s_0\) is the initial state of \(\D\). 
A trace \(\tau\) is \textit{legal}, if for every $i$ there exists an agent action \(a_i\) and an environment reaction \(r_i\) such that \(s_{i+1} = \delta(s_{i}, a_{i}, r_{i})\) with \(a_i \in \alpha(s_{i})\) and \(r_i \in \beta(s_{i}, a_i)\). 

We denote by $\vec{a} = a_0 a_1 \ldots$~(resp. $\vec{r} = r_0 r_1 \ldots$) a sequence of agent actions~(resp. environment reactions) and by \(\vec{a}^{\ k}\) (resp. \(\vec{r}^{\ k}\)) the prefix of \(\vec{a}\) (resp. \(\vec{r}\)~) up to action \(a_k\) (resp. reaction $r_k$).
We use \(\vec{a}_\epsilon\) and \(\vec{r}_\epsilon\) to denote the empty sequence of agent actions and environment reactions~(when $k < 0$), respectively. 
Consider $\aseq$ and $\rseq$ with the same length, we denote the trace induced by $\aseq$ and $\rseq$ on $\D$ as $\Trace(\aseq, \rseq)$ defined as follows: \begin{compactitem}
    \item $\Trace(\aseq_\epsilon, \rseq_\epsilon) = s_0$
    \item $\Trace(\aseq^{k}, \rseq^{k}) = \\     ~~~~\Trace(\aseq^{k-1}, \rseq^{k-1}) \cdot \delta(\last(\Trace(\aseq^{k-1}, \rseq^{k-1}), a_{k}, r_{k}))$
\end{compactitem}
If $\delta(\last(\Trace(\aseq^{k-1}, \rseq^{k-1}), a_{k}, r_{k}))$ is undefined or $\aseq$ and $\rseq$ are in different length, $\Trace(\aseq, \rseq)$ returns undefined.

An agent strategy is a function $\sigma: (2^\F)^+ \total \act$ mapping traces of states on $\D$ to agent actions. Note that in the synthesis literature, agent strategies are typically defined as functions $\sigma': \react^* \total \act$ mapping histories of environment reactions to agent actions. However, we define agent strategies equivalently as functions $\sigma: (2^\F)^+ \total \act$ to be more in line with existing works on planning and reasoning about actions~\cite{DeGR18}. The equivalence follows by noting that: \myi every strategy $\sigma: (2^\F)^+ \total \act$ directly corresponds to a strategy $\sigma': \react^* \total \act$ by the requirement of uniqueness of environment reaction; \myii every strategy $\sigma': \react^* \total \act$ directly corresponds to a strategy $\sigma: (2^\F)^+ \total \act$ since the transition function $\delta$ is deterministic. An agent strategy $\sigma$ is \textit{legal} if, for every legal trace \(\tau\), it holds that for all its prefixes \(\tau^k\), if \(a_{k} = \sigma(\tau^k)\), then \(a_{k} \in \alpha(\last(\tau^k))\).

An \textit{environment strategy} is a function $\gamma: Act^+ \rightarrow React$.
Given a sequence of agent actions \(\vec{a}\) and an environment strategy $\gamma$, we denote by \(\vec{\gamma}(\vec{a})\) the sequence of environment reactions obtained by applying \(\gamma\) recursively on every finite prefix \(\vec{a}^{\ k}~(k \geq 0)\) of \(\vec{a}\), i.e., \(\vec{\gamma}(\vec{a})_k = \gamma(\vec{a}^{\ k})\). 
An environment strategy is \textit{legal}, if for every 
sequence of agent actions $\vec{a}$, it holds that for all its prefixes $\vec{a}^{\ k}$, if \(\tau = \mathtt{Trace}(\vec{a}^{k-1}, \vec{\gamma}(\vec{a}^{k-1}))\) is defined, hence legal, and \({a}_k \in \alpha(\last(\tau))\), then \(r_{k} = \gamma(\vec{a}^{\ k-1} \cdot a_k) \in \beta(\last(\tau), a_k))\).

Given a legal agent strategy \(\sigma\) and a legal environment strategy \(\gamma\), there exists a unique trace of \(\D\) that is \textit{induced} by both \(\sigma\) and \(\gamma\), which we denote as \(\mathtt{Play}(\sigma, \gamma)\). Formally, \(\mathtt{Play}(\sigma, \gamma) = s_0 s_1 \ldots \) is such that \(s_0\) is the initial state of \(\D\) and for every $i \geq 0$, \(s_{i+1} = \delta(s_i, a_{i}, r_{i})\) with \(a_{i} = \sigma(s_0 \ldots s_i)\) and \(r_{i} = \gamma(a_0 \ldots a_{i})\). 
We can prove by induction that $\mathtt{Play}(\sigma, \gamma)$
is indeed defined and \textit{legal}. 

\begin{theorem}~\label{thm:legality}
Let \(\D\) be a planning domain, \(\sigma\) a legal agent strategy, and \(\gamma\) a legal environment strategy. Then, \(\mathtt{Play}(\sigma, \gamma)\) is defined and legal.
\end{theorem}

\noindent\textbf{Strong and Cooperative Solutions.}
Given a planning domain $\D$, an agent \emph{goal} is an \LTLf formula $\varphi$ defined over the fluents $\F$ of the domain $\D$. A legal trace $\ptrace$ of $\D$ satisfies $\varphi$ in $\D$, written $\tau \pmodels \varphi$, if there exists a prefix of $\ptrace$ that satisfies $\varphi$. An agent strategy $\sigma$ is a \emph{strong solution} for $\varphi$ in $\D$ if, $\sigma$ is legal and for every legal environment strategy $\gamma$, it holds that $\Play(\sigma, \gamma) \pmodels \varphi$. If such a strategy $\sigma$ exists, we say that there exists a strong solution for $\varphi$ in $\D$, which is $\sigma$. Furthermore, an agent strategy $\sigma$ is a \emph{cooperative solution} for $\varphi$ in $\D$, if $\sigma$ is legal and there exists a legal environment strategy $\gamma$ such that $\Play(\sigma, \gamma) \pmodels \varphi$. If such a strategy $\sigma$ exists, there exists a cooperative solution for $\varphi$ in $\D$, which is $\sigma$.

\subsection{Best-Effort Synthesis in Planning Domains~\label{sec:3-2}}

We now introduce the problem of \LTLf best-effort synthesis in nondeterministic planning domains. In doing so, we need to define what it means for an agent to make its best-effort to achieve its goal in a planning domain. To formalize such strategies, we adapt the definitions in~\cite{ADR2021} and consider first the notion of dominance.


\begin{definition}~\label{dfn:dominance}
Let \(\D\) be a planning domain, \(\varphi\) an \LTLf formula, and \(\sigma_1, \sigma_2\) two legal agent strategies. \(\sigma_1\) \textit{dominates} \(\sigma_2\) for \(\varphi\) in \(\D\), written \(\sigma_1 \geq_{\varphi|\D} \sigma_2\), if for every legal environment strategy \(\gamma\), \(\mathtt{Play}(\sigma_2, \gamma) \models_{\D} \varphi\) implies \(\mathtt{Play}(\sigma_1, \gamma) \models_{\D }\varphi\).
\end{definition}

Furthermore, a legal agent strategy \(\sigma_{1}\) \textit{strictly dominates} legal agent strategy \(\sigma_{2}\) for goal \(\varphi\) in \(\D\), written \(\sigma_1 >_{\varphi|\D} \sigma_2\), if \(\sigma_1 \geq_{\varphi|\D} \sigma_2\) and \(\sigma_2 \not \geq_{\varphi|\D} \sigma_1\). Intuitively, \(\sigma_1 >_{\varphi|\D} \sigma_2\) shows that \(\sigma_1\) does at least as well as \(\sigma_2\) against every legal environment strategy in \(\D\) and strictly better against at least one such strategy. If $\sigma_1$ strictly dominates $\sigma_2$, then an agent using $\sigma_2$ is not doing its “best” to achieve the goal.
Within this framework, a best-effort solution is a legal agent strategy that is not strictly dominated by any other legal strategies.

\begin{definition}~\label{dfn:best-effort}
A legal agent strategy $\sigma$ is a best-effort solution for $\varphi$ in $\D$ iff there is no legal agent strategy $\sigma'$ such that $\sigma' >_{\varphi|\D} \sigma$.
\end{definition}

It is worth noting that, although we adapt the definitions of best-effort solutions from~\cite{ADR2021}, the key difference is that, in our framework, best-effort solutions are  required to be legal agent strategies, i.e., to always satisfy agent action preconditions. However, in the best-effort synthesis setting defined in~\cite{ADR2021}, since the environment is specified as an \LTL/\LTLf formula, there are no such preconditions for the agent to follow.

\begin{definition}~\label{def:best-effort-synthesis}
The problem of \LTLf best-effort synthesis in nondeterministic planning domains is defined as a pair $\P = (\D, \varphi)$, where $\D$ is a nondeterministic planning domain and $\varphi$ is an \LTLf formula. Best-effort synthesis of $\P$ computes a best-effort solution for $\varphi$ in $\D$.
\end{definition}

In particular, if the synthesis problem $\P = (\D, \varphi)$ admits a strong solution for $\varphi$ in $\D$, then computing a best-effort solution coincides with computing a strong solution. This observation is analogous to the one for \LTLf best-effort synthesis in~\cite{ADR2021}, where the environment is specified in \LTL/\LTLf.


\begin{proposition}
Let $\P = (\D, \varphi)$ be the defined synthesis problem that admits a strong solution for $\varphi$ in $\D$ and $\sigma$ an agent strategy. We have that 
$\sigma$ is a best-effort solution for $\varphi$ in $\D$ iff $\sigma$ is a strong solution for $\varphi$ in $\D$.
\end{proposition}

In order to compute best-effort solutions, we first observe that a best-effort solution can be considered as a strategy $\sigma$ such that for every history $h \in (2^\F)^+$, i.e., a finite and legal sequence of states of a planning domain $\D$, that is consistent with $\sigma$, 
$\sigma$ extends $h$ to fulfill the goal whenever possible. Hence, one of the following possibilities holds: \myi if the agent can extend $h$ to fulfill the goal regardless of how the environment reacts, then any extension of $h$ following $\sigma$ should ensure goal completion; \myii if no extension of $h$ allows the agent to fulfill the goal, then $\sigma$ just behaves legally;
\myiii if there are only some extensions of $h$, but not all, that allow the agent to fulfill the goal, then there should exist an extension of $h$ following $\sigma$ that fulfills the goal. Observe that \myiii guarantees that the agent does its best, assuming the environment might cooperate to help the agent fulfill its goal. 

Consider an agent strategy $\sigma$. We now formalize these three possibilities by assigning a value to each history $h$ consistent with $\sigma$, as in~\cite{ADR2021}. 
%
%
For a legal agent strategy $\sigma$ and a history $h$ that is consistent with $\sigma$, we use $\Gamma(h, \sigma)$ to denote the set of legal environment strategies $\gamma$ in $\D$ such that $h$ is a prefix of $\Play(\sigma, \gamma)$. Furthermore, we denote by $\H_{\D}(\sigma)$ the set of legal histories $h$ such that $\Gamma(h, \sigma)$ is nonempty, i.e., the set of legal histories of $\D$ that are consistent with $\sigma$ with some legal environment strategy $\gamma$. We define the value of $h$ that is consistent with a legal agent strategy $\sigma$ as follows:

\begin{compactitem}
    \item $\val(\sigma, h) = +1$ (winning), if $\Play(\sigma, \gamma) \pmodels \varphi$ for every $\gamma \in \Gamma(h, \sigma)$;
    \item $\val(\sigma, h) = -1$ (losing), if $\Play(\sigma, \gamma) \not \pmodels \varphi$ for every $\gamma \in \Gamma(h, \sigma)$;
    \item $\val(\sigma, h) = 0$ (pending), otherwise.
\end{compactitem}

Let $\val(h)$ be the maximal value of $\val(\sigma, h)$, considering all the strategies $\sigma$ such that $h \in \H_{\D}(\sigma)$\footnote{We consider $\val(h)$ only if there exists at least one agent strategy $\sigma$ such that $h \in \H_{\D}(\sigma)$.}. The following theorem states that a best-effort solution $\sigma$ guarantees to maximize the value of every history $h$ that is consistent with $\sigma$.

\begin{theorem} [Maximality Condition]\label{thm:max}
An agent strategy $\sigma$ is a best-effort solution for $\varphi$ in $\D$ iff for every $h \in \H_{\D}(\sigma)$ it holds that $\val(\sigma, h) = \val(h)$.
\end{theorem}

While classical synthesis and planning settings (see e.g.~\cite{PR89,DegVa15,DeGR18}) first require checking the realizability of the problem, i.e., the existence of a solution, in the case of best-effort synthesis realizability is trivial, as there always exists a best-effort solution.

\begin{theorem}~\label{thm:existence}
Let $\P = (\D, \varphi)$ be a problem of \LTLf best-effort synthesis in nondeterministic planning domains. There always exists a best-effort solution $\sigma$ for $\varphi$ in $\D$. 
\end{theorem}

In order to prove Theorem~\ref{thm:existence}, we show that we can always construct a strategy $\hat{\sigma}$ that satisfies the Maximality Condition. To do so, we construct a chain of legal agent strategies $\sigma_0, \sigma_1, \ldots$ by selecting for each $i \geq 0$, a history $h \in \H_\D(\sigma_i)$ that has not yet been marked as stabilized, i.e., such that $val(\sigma_i, h) < val(h)$, and maximizing the value $val(\sigma_{i+1}, h)$. $\sigma_{i+1}$ is constructed by selecting a strategy $\sigma$ such that $val(\sigma, h) = val(h)$ and setting $\sigma_{i+1} = \sigma_i[h \leftarrow \sigma]$, where $\sigma_i[h \leftarrow \sigma]$ denotes the strategy that agrees with $\sigma_i$ everywhere, except in $h$ and all its extensions where it agrees with $\sigma$. $\hat{\sigma}$ is the point-wise limit of the chain of strategies, i.e., $\hat{\sigma} = lim_{i} \ \sigma_{i}$.

\section{Synthesizing Best-Effort Solutions~\label{sec:4}}

We now provide an algorithm for \LTLf best-effort synthesis in nondeterministic planning domains based on game-theoretic techniques. In particular, we use \DFA games for adversarial/cooperative reachability and safety~\cite{DegVa15,ADR2021,ZhuTLPV17} to capture the environment being adversarial/cooperative and agent always following its action preconditions, respectively.  
%
%
\subsection{\DFA Games~\label{sec:4-1}}


A \emph{\DFA game} is a pair $\A = (\T, \acc)$, where $\T$ is a deterministic transition system acting as the \emph{game arena} and $Acc \subseteq Q^\omega$ is the acceptance condition. Specifically, $\T = (Act \times React, Q, q_0, \varrho)$ is a \textit{deterministic transition system}, where $Act \times React$ is the alphabet, in which $\act$ and $\react$ are two disjoint sets of variables under the control of the agent and the environment, respectively;  $Q$ is a finite set of states; $q_0 \in Q$ is the initial state; $\varrho\colon Q \times Act \times React \rightarrow Q$ is the transition function. 
Given an infinite word $\pi = \pi_0 \pi_1 \ldots \in (Act \times React)^\omega$, the \textit{run} $\rho=\run(\pi, \T)$ of $\pi$ on $\T$ is an infinite sequence $\rho = q_0 q_1 \ldots \in Q^{\omega}$ such that $q_0$ is the initial state of $\T$ and $q_{i+1} = \varrho(q_i, \pi_i)$ for every $i \geq 0$. 
Analogously, we denote by $\rho^{k+1} = \run(\pi^k, \T)$ the finite sequence $\rho^{k+1} = q_0q_1\ldots q_{k+1}$ obtained from running $\pi^k$ on $\T$. 
A run $\rho$ is \emph{accepting} if $\rho \in Acc$. We denote by $\L(\A)$ the set of words $\pi$ accepted by $\A$, i.e., such that $\rho = \run(\pi, \T)$ is accepting. 
In this work, we specifically consider the following acceptance conditions:

\begin{compactitem}
    \item \textit{Reachability}. Given a set $R \subseteq Q, \reach(R) = \{q_0 q_1 \ldots \in Q^\omega \mid  \exists k \geq 0.q_k \in R\}$, i.e., a state in $R$ is visited at least once.
    \item \textit{Safety}. Given a set $S \subseteq Q, \safe(S) = \{q_0 q_1 \ldots \in Q^\omega \mid  \forall k \geq 0.q_k \in S\}$, i.e., only states in $S$ are visited.
    \item \textit{Safety-Reachability}. Given two sets $S, R \subseteq Q, \safereach(S, R) $ 
    $=\{q_0 q_1 \ldots \in Q^\omega \mid \exists \ell \geq 0. q_\ell  \in R \text{ and } \forall 0 \leq k \leq \ell.  q_k \in S\}$, i.e., a state in $R$ is visited at least once and until then only states in $S$ are visited.
\end{compactitem}

Notably, a deterministic transition system with reachability acceptance condition defines a deterministic finite state automaton~(\DFA).

An \textit{agent strategy} 
on a \DFA game is a function $\kappa: Q \total \act$.\footnote{For the \DFA games considered in this paper, it is sufficient to define strategies in this form, which are called \emph{positional strategies}. More general forms of strategies, mapping histories of the game to agent actions/environment reactions, are only needed for more sophisticated games~\cite{games}.}
Given an agent strategy $\kappa$, a sequence of environment reactions $\vec{r} = r_0 r_1 \ldots \in \react^\omega$, and a transition system $\T$, we denote by $\run(\kappa, \vec{r}, \T)$, the unique sequence $q_0 q_1 \ldots \in Q^\omega$ on $\T$ \emph{induced} by~(or \emph{consistent with}) $\kappa$ and $\vec{r}$ as follows: $q_0$ is the initial state of $\T$, and for every $i \geq 0$, $q_{i+1} = \varrho(q_{i}, a_{i}, r_{i})$, where $a_i = \kappa(q_i)$.

An agent strategy $\kappa$ is \emph{winning} in the game $\A = (\T, \acc)$, if for every sequence of environment reactions $\vec{r} \in \react^\omega$, it holds that $\run(\kappa, \vec{r}, \T) \in \acc$.
In \emph{\DFA games}, $q \in Q$ is a \emph{winning} state, if the agent has a winning strategy in the game $\A' = (\T', Acc)$, where $\T'=(Act \times React, Q, q, \delta)$, i.e., on the same structure but with a new initial state $q$. We denote by $W_{adv}(\T, Acc)$ (sometimes $W_{adv}$ when $\T$ and $Acc$ are clear from the context) the set of all agent winning states. Intuitively, $W_{adv}$ represents the ``agent winning region", from which the agent can win the game, no matter how the environment behaves. A positional strategy that is winning from every state in the winning region is called \emph{uniform winning}.

Similarly, an agent strategy $\kappa$ is \emph{cooperatively winning} in a game $\A = (\T, \acc)$ if there exists a sequence of environment reactions $\vec{r} \in \react^\omega$ such that $\run(\kappa, \vec{r}, \T) \in \acc$.
Hence, \(q \in Q\) is a \emph{cooperatively winning state}, if the agent has a cooperatively winning strategy in the game \(\A' = (\T', Acc)\), where \(\T' = (Act \times React, Q, q, \delta)\).  By 
$W_{coop}(\T, Acc)$ (sometimes $W_{coop}$), we denote the set of all the agent cooperatively winning states. A positional strategy that is winning from every state in the cooperatively winning region is called \emph{uniform cooperatively winning}.

\subsection{Synthesis Technique~\label{sec:4-2}}
The key idea of our game-theoretic synthesis approach is reducing the synthesis problem to  DFA games, where the game arena is obtained by suitably composing the nondeterministic planning domain and the DFA of the \LTLf formula. We first show how to transform the nondeterministic planning domain $\D$ into a deterministic transition system, which is then composed with the DFA of the \LTLf formula to obtain the game arena. 
%

In order to transform the domain $\D$ to a deterministic transition system, we need to drop the preconditions of both agent actions and environment reactions and transform the partial transition function into a total transition function.
To that end, we introduce two new states, \(s^{ag}_{err}\) and \(s^{env}_{err}\), indicating that the agent and the environment violate their respective preconditions. Formally, given a nondeterministic planning domain $\D = (2^\F, s_0, Act, React, \alpha, \beta, \delta)$, the corresponding deterministic transition system \(\D_+ = (Act \times React, 2^\F \cup \{s^{ag}_{err}, s^{env}_{err}\}, s_0, \delta'\)) is constructed as follows:
\begin{compactitem}
    \item $Act \times React$ is the alphabet; 
    \item 
    $2^\F \cup \{s^{ag}_{err} \cup s^{env}_{err}\}$ is the state space, where $s^{ag}_{err}$ is the agent error state, and $s^{env}_{err}$ is the environment error state;
    \item
    \(s_0 \in 2^\F\) is the initial state;
    \item
    \(\delta': (2^\F \cup \{s^{ag}_{err}, s^{env}_{err}\}) \times Act \times React \rightarrow 2^\F \cup \{s^{ag}_{err}, s^{env}_{err}\}\) is such that if $s \in \{s^{ag}_{err}, s^{env}_{err}\}$ then $\delta'(s, a, r) = s$, otherwise
    $$
    \delta'(s, a, r) = 
    \begin{cases}
    \delta(s, a, r) &\text{ if } a \in \alpha(s) \text{ and } r \in \beta(s, a) \\
    s^{ag}_{err} &\text{ if } a \not \in \alpha(s) \\
    s^{env}_{err} &\text{ if } a \in \alpha(s) \text{ and } r \notin \beta(s, a).
    \end{cases}
    $$

\end{compactitem} 

Intuitively, the transitions of $\D_+$ are the same as in $\D$, except that the transitions move to $\agerr$ and $\enverr$ when agent action or environment reaction preconditions are violated, respectively. Once an error state is reached, $\D_+$ just keeps looping there.

We now construct the game arena, on which the agent and the environment control actions and reactions, respectively, through an ad-hoc composition of $\D_+$ and (the transition system of) the \DFA of the agent goal $\varphi$.
%
Given an \LTLf formula \(\varphi\) over \(\F\), we first obtain its corresponding \DFA \(\A_{\varphi} = (\T_{\varphi}, \reach(R_{\varphi}))\), where \(\T_{\varphi} = (2^\F, Q, q_0, \varrho)\) is the transition system and \(R_{\varphi} \subseteq Q\) is the set of final states. 
Note that the transitions of $\A_{\varphi}$ are defined wrt fluents $\F$, i.e., the transition function of $\T_{\varphi}$ is in the form of \(\varrho: Q \times 2^\F \rightarrow Q\). But the transitions of $\D_+$ are defined wrt agent actions and environment reactions.
Hence, the composition of $\D_+$ and $\T_{\varphi}$ needs to suitably synchronize the transitions of both. To that end, we map the fluent evaluations in the states of \(\D_+\) to the transition conditions that follow the transition function \(\varrho\).

Formally, the composition $\T = \D_+ \circ \T_{\varphi}$ is such that $\T = (Act \times React, (2^\F \cup \{s^{ag}_{err}, s^{env}_{err}\}) \times Q, (s_0, \varrho(q_0,s_0)), \partial)$, where:

\begin{compactitem}
    \item \(Act \times React\) is the alphabet;
    \item \((2^\F \cup \{s^{ag}_{err}, s^{env}_{err}\}) \times Q\) is the set of states;
    \item \((s_0, \varrho(q_0,s_0))\) is the initial state;
    \item \(\partial: ((2^\F \cup \{s^{ag}_{err}, s^{env}_{err}\}) \times Q) \times Act \times React \rightarrow ((2^\F \cup \{s^{ag}_{err}, s^{env}_{err}\}) \times Q)\) is the transition function such that: 
     $$
     \partial((s, q), a, r) =
      \begin{cases}
    (s', \varrho(q, s')) &\text{if } s' \not \in \{s^{ag}_{err}, s^{env}_{err}\} \\
    (s^{ag}_{err}, q) &\text{if }  s' = s^{ag}_{err}\\
    (s^{env}_{err}, q) &\text{if } s' = s^{env}_{err}
    \end{cases}
    $$
    where $s' = \delta'(s, a, r)$.
    
\end{compactitem}


Intuitively, $\T$ is a deterministic transition system that simultaneously retains the state of the domain $\D$ and the progress in the DFA $\A_{\varphi}$ in satisfying the \LTLf goal $\varphi$. In particular, note that the function \(\partial\) is defined such that if the $\T$ reaches an environment/agent error state, $\T$ cannot reach any other state afterwards.
%
%
In fact, a positional strategy $\kappa$ on $\T = \D_+ \circ \T_{\varphi}$ also induces a strategy $\sigma': \react^* \total \act$ in the planning domain $\D$.

\begin{definition}~\label{dfn:induced-strategy}
Let $\kappa: Q \total \act$ be a positional strategy on $\T = \D_+ \circ \T_{\varphi}$. $\kappa$ induces an agent strategy $\sigma': \react^* \total \act$: $\sigma'(\rseq_{\epsilon}) = \kappa(t_0)$, where $t_0$ is the initial state of $\T$; for every $i \geq 0$, \(\sigma'(\rseq^{\ i}) = \kappa(t)\), where $\rseq^{\ i} = r_0 \ldots r_i$ and $t$ is the last state in the finite sequence $\rho^{i+1} = \run(\pi^i, \T)$ with $\pi^i = (\sigma'(\rseq_{\epsilon}) \cup r_0)(\sigma'(\rseq^{\ 0}) \cup r_1)\ldots(\sigma'(\rseq^{\ i-1}) \cup r_i)$.
\end{definition}

Indeed, with $\sigma'$ we can also obtain the equivalent strategy $\sigma: (2^\F)^+ \total \act$ (c.f. Section~\ref{sec:3-1}) in $\D$. Sometimes, for simplicity, we directly say that positional strategy $\kappa$ induces strategy $\sigma: (2^\F)^+ \total \act$ (rather than saying that $\kappa$ induces strategy $\sigma': \react^* \total \act$ which is equivalent to a strategy $\sigma: (2^\F)^+ \total \act$).


Given a synthesis problem $\P=(\D, \varphi)$, we now detail how to compute a best-effort solution for $\varphi$ in $\D$ by reducing to suitable DFA games on the game arena $\T = (Act \times React, (2^\F \cup \{s^{ag}_{err}, s^{env}_{err}\}) \times Q, (s_0, \varrho(q_0,s_0)), \partial)$ constructed as above. 
In the following, we denote by $R'_{\varphi}$ the set of states in $\T$ by lifting the final states $R_{\varphi}$ in $\T_{\varphi}$ to $\T$. Hence, $R'_{\varphi} = \{ (s,q) \mid q \in R_{\varphi}\}$. Moreover, we denote by $\bigagerr$ and $\bigenverr$ the sets of states by lifting the agent error state and the environment error state in $\D_+$ to $\T$, respectively. Therefore, $\bigagerr$ and $\bigenverr$ indicate the agent and the environment violate their respective preconditions. Hence $\bigagerr = \{(s, q) \mid s = s^{ag}_{err}\}$ and $\bigenverr = \{(s, q) \mid s = s^{env}_{err}\}$). Sometimes, we write $\neg \bigagerr$ as an abbreviation for $((2^\F \cup \{\agerr, \enverr\}) \times Q) \setminus \bigagerr$ to indicate that the agent has not yet violated its precondition. Similarly for $\neg \bigenverr$.




\medskip
\noindent \textbf{Algorithm 1.} Given problem $\P = (\D, \varphi)$ of \LTLf best-effort synthesis in nondeterministic planning domains, proceed as follows:
 \begin{compactenum}
 \item 
 Construct the \DFA \(\A_{\varphi} = (\T_{\varphi}, \reach(R_{\varphi}))\) of agent goal $\varphi$ and the transition system \(\D_+\) of domain \(\D\).
 \item 
 Construct arena \(\T = (\D_+ \circ \T_{\varphi})\) by composition.
 \item 
 In the \DFA game $(\T, \reach(\lnot S^{ag}_{err} \cap ( S^{env}_{err} \cup R'_\varphi)))$, compute a positional uniform winning strategy $\kappa_{adv}$. Let $W_{adv}$ be the winning region.
 \item 
 In the \DFA game $(\T, \reach(\lnot S^{ag}_{err} \cap \lnot S^{env}_{err} \cap R'_\varphi))$, compute a positional uniform cooperatively winning strategy $\kappa_{coop}$. Let $W_{coop}$ be the cooperatively winning region. 
 \item
 Construct positional strategy $\kappa$ from $\kappa_{adv}$ and $\kappa_{coop}$ as follows:
    $$
    \kappa(s, q) =
    \begin{cases}
        \kappa_{adv}(s, q) &\text{ if } (s, q) \in W_{adv} \\
        \kappa_{coop}(s, q) &\text{ if } (s, q) \in W_{coop} \backslash W_{adv} \\
        \text{any } a \in \alpha(s) &\text{ otherwise}
    \end{cases}
    $$
    where $\alpha(s) = Act$, if $s \in \{s^{ag}_{err}, s^{env}_{err}\}$, such that the agent can perform any action at an error state.
    \item \textbf{Return} the (best-effort) solution $\sigma$ for $\varphi$ in $\D$ \textit{induced} by $\kappa$.

\end{compactenum}


Due to the requirement of \emph{existence of agent action}, at every state $s \in 2^\F$, there always exists an agent action $a \in \alpha(s)$~(possibly $nop$). 

It is worth noting that Algorithm~1 also allows us to check whether the computed best-effort solution is a strong solution as well. This is indicated by the value of $val(\sigma, s_0)$. More specifically, we have that the computed solution $\sigma$ is a strong solution if $t_0 \in W_{adv}$, where $t_0$ is the initial state of $\T = \D_+ \circ \T_{\varphi}$.  


\begin{theorem}[Correctness]~\label{thm:correctness}
    Let $\D$ be a nondeterministic planning domain, $\varphi$ an \LTLf goal, and $\sigma$ the agent strategy returned by Algorithm~1. Then, $\sigma$ is a best-effort solution for $\varphi$ in $\D$. 
\end{theorem}

To prove this theorem, we make use of several intermediate results, which are given in the next section.

\begin{figure*}[th]
     \centering
     \includegraphics[width=\linewidth]{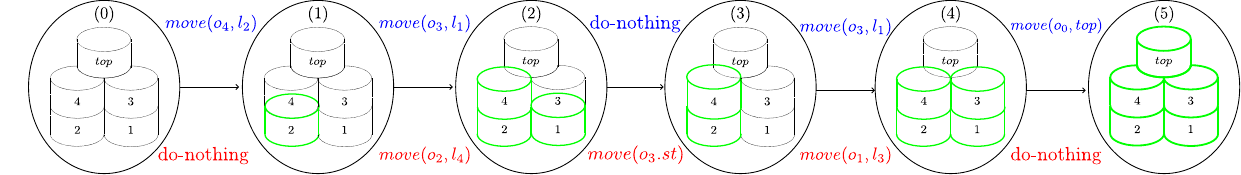}
         \caption{An execution example of a best-effort solution for the arch-building task described in Section~\ref{sec:1-1}. Robot actions and human reactions  are shown in blue and red, respectively.}
    \label{fig:validation}
\end{figure*}
\subsection{Correctness~\label{sec:4-3}}

In order to prove Theorem~\ref{thm:correctness}, we first show the connection between strong and cooperative solutions for the agent goal in a planning domain to winning and cooperatively winning strategies in the corresponding \DFA games, respectively. 

\begin{theorem}~\label{thm:sol-win}
Let $\D$ be a planning domain, $\varphi$ an \LTLf goal, and $\T = \D_+ \circ \T_{\varphi}$ the constructed game arena. The following two hold: \begin{compactenum}
    \item[\mya] 
    there is a strong solution for $\varphi$ in $\D$ iff the agent has a winning strategy in $(\T, \safereach(\lnot S^{ag}_{err},\, S^{env}_{err} \cup R'_\varphi))$.
    \item[\myb] 
    there is a cooperative solution for $\varphi$ in $\D$ iff the agent has a cooperatively winning strategy in $(\T, \safereach(\lnot S^{ag}_{err} \cap \lnot S^{env}_{err}, R'_\varphi))$.
\end{compactenum}
\end{theorem}

To prove Theorem~\ref{thm:sol-win}, 
%
%
we first observe that a strong solution for $\varphi$ in $\D$ guarantees that the agent always follows its action precondition~(i.e., never visits $S^{ag}_{err}$) \textit{and} either forces the environment to violate its reaction precondition or eventually satisfies $\varphi$~(i.e., eventually reaches $S^{env}_{err} \cup R'_{\varphi}$ ). 
Hence, the agent's goal is to satisfy {\centerline{$\Box(\lnot S^{ag}_{err})\land \Diamond(S^{env}_{err}\lor R'_\varphi)$}}
(For simplicity, we abuse notations and consider sets of states as atomic propositions). Therefore, the reduced game for computing a strong solution is $(\T, \safereach(\lnot S^{ag}_{err},\, S^{env}_{err} \cup R'_\varphi))$. Similarly for proving \myb.

Next, we show that
the safety-reachability games can be solved by reducing to pure reachability games, the correctness of which can be shown by construction.

\begin{lemma}~\label{lem:safe-to-reach}
Let \(\T\), $S^{ag}_{err}$, $S^{env}_{err}$ and $R'_{\varphi}$ be as defined above. Then, for every run $\rho$ on $\T$: \begin{compactitem}
    \item 
    $\rho$ is an accepting run of $\safereach(\lnot S^{ag}_{err},\, S^{env}_{err} \cup R'_\varphi)$ iff $\rho$ is an accepting run of $\reach(\lnot S^{ag}_{err}\cap( S^{env}_{err}\cup R'_\varphi))$.
    \item 
    $\rho$ is an accepting run of $\safereach(\lnot S^{ag}_{err} \cap \neg S^{env}_{err}, R'_\varphi)$ iff $\rho$ is an accepting run of $\reach(\lnot S^{ag}_{err} \cap S^{env}_{err} \cap R'_\varphi)$.
\end{compactitem}
\end{lemma}


%
By Theorem~\ref{thm:sol-win} and Lemma~\ref{lem:safe-to-reach}, the computed winning strategy $\kappa_{adv}$ for $\reach(\lnot S^{ag}_{err}\cap( S^{env}_{err}\cup R'_\varphi))$ and the cooperative winning strategy $\kappa_{coop}$ for $\reach(\lnot S^{ag}_{err} \cap S^{env}_{err} \cap R'_\varphi)$ indeed correspond to a strong solution and a cooperative solution for $\varphi$ in $\D$, respectively. 

We now show that the computed final strategy $\sigma$ by combing $\kappa_{adv}$ and $\kappa_{coop}$ is a best-effort solution for $\varphi$ in $\D$. We prove this through Theorem~\ref{thm:max} by showing that $\sigma$ satisfies the Maximality Condition.

\begin{theorem}\label{thm:local-correct}
Let $\sigma$ be the strategy returned by Algorithm 1. For every $h \in \H_\D(\sigma)$ it holds $val(\sigma, h) = val(h)$.
\end{theorem}

To prove Theorem~\ref{thm:local-correct}, first observe that every sequence of states $\rho$ on $\T$ that does not end at an environment/agent error state corresponds to a legal history $h$ on $\D$. The strategy $\sigma$ returned by Algorithm~1 satisfies that, for every $h \in \H_\D(\sigma)$, one of the following holds:
\myi the corresponding run $\rho$ on $\T$ ends at a state $s \in W_{adv}$ hence $h$ can surely be extended to satisfy $\varphi$ in $\D$ using $\sigma$ (i.e., from $h$ the agent has a strong solution); \myii the corresponding run $\rho$ on $\T$ ends at a state $s \in W_{coop} \backslash W_{adv}$ hence $h$ can possibly be extended to satisfy $\varphi$ in $\D$ using $\sigma$, should the environment cooperate (i.e., from $h$ the agent has a cooperative solution); \myiii the corresponding run $\rho$ on $\T$ ends at a state $s \notin W_{coop} \cup W_{adv}$ hence $h$ cannot be extended to satisfy $\varphi$ in $\D$ (i.e., from $h$ the agent has neither a strong nor a cooperative solution). Moreover, $\sigma$ is guaranteed to be a legal strategy since the agent can only perform legal actions. 

Finally, we can prove Theorem~\ref{thm:correctness}~(the correctness of Algorithm 1 in Section~\ref{sec:4-2}) as an immediate result of Theorems~\ref{thm:local-correct} and~\ref{thm:max}.


\subsection{Computational Complexity~\label{sec:4-4}}


The computational complexity of \LTLf best-effort synthesis in nondeterministic planning domains is the following:
\begin{theorem}~\label{thm:complexity}
\LTLf best-effort synthesis in nondeterministic planning domains is:
\begin{compactitem}
    \item 2EXPTIME-complete in the size of the \LTLf goal;
    \item EXPTIME-complete in the size of the domain.
\end{compactitem}
\end{theorem}


Regarding the complexity in the size of the \LTLf goal, the hardness comes from \LTLf synthesis, and the membership comes from the \LTLf-to-\DFA construction~(double-exponential in the size of the \LTLf formula). Hence, if we consider simple reachability goals in the form of $\varphi = \Diamond(R)$, where $R$ is a propositional formula over the fluents $\F$ of the planning domain, the problem is just polynomial in the size of $\varphi = \Diamond(R)$ since the \DFA can be constructed in polynomial time. Regarding the complexity in the size of the domain, the hardness comes from planning itself \cite{DeGR18}, and the membership comes from the construction of the deterministic transition system from the planning domain~(single-exponential in the number of fluents).

An interesting observation from Theorem~\ref{thm:complexity} is that best-effort synthesis provides an efficient alternative approach to planning in nondeterministic domains for both \LTLf and reachability goals. In fact, instead of looking for a strong solution, which may not exist, one can look for a best-effort solution, which, on the one hand, always exists and on the other hand is directly a strong solution if the problem admits one.

\noindent\textbf{Running Example (cont.).~\label{sec:5-1}}
We present the computed best-effort solution produced by our synthesis algorithm for the running example in Section~\ref{sec:1-1}.
We represent the arch-building task as a suitable \LTLf formula and the dynamics of the interactions between the robot and the human (seen as the agent and the environment, respectively) as a nondeterministic planning domain.
Figure~\ref{fig:validation} presents a (simplified) execution example of a best-effort solution to building the arch. 
Note that at the initial state, all blocks are placed in storage~(state $0$). Then the robot successfully placed the block $o_4$ at the location $l_2$ without any interference from the human~(state $1$). Next, the robot placed the block $o_3$ at $l_1$, and the human helped build the arch by placing $o_2$ at $l_4$~(state $2$). Seeing the human being cooperative, the robot decides to be lazy, hence not doing anything. However, the human now punishes the robot for being lazy by undoing what the robot has done, thus removing $o_3$ back to storage (state $3$). The robot now realizes that the human is not always cooperative, so it proceeds by placing $o_3$ back at $l_1$, and the human cooperates by placing $o_1$ at $l_3$~(state $4$). Finally, the robot can build the arch by placing $o_0$ at the top, with no interference from the human~(state $5$).
This example well illustrates the basic characteristic of best-effort solutions: handling both adversarial and cooperative environment behaviors, even if the environment switches back and forth between behaving adversarially and cooperatively instead of always being adversarial or cooperative.


\section{Implementation and Empirical Evaluation}
We implemented our algorithm for \LTLf best-effort synthesis in planning domains 
in a tool called \textit{BeSyftP}, leveraging the symbolic \LTLf synthesis framework~\cite{ZTLPV17} that is integrated in all state-of-the-art \LTLf
synthesis tools~\cite{BLTV,DeGiacomoF21}. The explicit-state \DFAs of \LTLf formulas are constructed by \lydia~\cite{lydia}, the overall best-performing tool for \LTLf-to-\DFA construction. \textit{BeSyftP} uses the \emph{symbolic} \DFA encoding proposed in~\cite{ZTLPV17} to represent the \DFAs symbolically and integrates the symbolic encoding of a planning domain~(specified in a variant of PDDL) proposed in~\cite{KWKLV19} for symbolic domain representation. Both symbolic \DFAs and planning domains are represented in
Binary Decision Diagrams~(BDDs)~\cite{Bryant92}, with the BDD library CUDD-3.0.0~\cite{cudd}. Following~\cite{ZTLPV17}, \textit{BeSyftP} constructs and solves symbolic \DFA games using Boolean operations provided by CUDD, such as quantification, negation and conjunction. In particular, positional strategies are obtained with Boolean synthesis~\cite{FriedTV16}. 
Finally, best-effort strategies are computed by applying suitable Boolean operations to the obtained positional winning strategy and cooperatively winning strategy.

%
%
\noindent\textbf{Experimental Comparison.} 
To show the efficiency of our technique for \LTLf best-effort synthesis in nondeterministic planning domains, we want to compare it with the approach of reducing to \LTLf best-effort synthesis, by re-expressing the domain as an \LTLf formula. More specifically, we adapted the domain-to-\LTLf translation in~\cite{KWKLV19} to obtain the \LTLf formula $\E$ of the nondeterministic planning domain. Next, we can utilize the \LTLf best-effort synthesis technique from~\cite{ADR2021} to solve the problem of $\varphi$~(agent goal) under $\E$. 

We also evaluated the performance of \besyftp considering the overhead of computing best-effort solutions wrt computing strong and cooperative solutions. Therefore, we also implemented the adversarial and cooperative synthesis approaches for computing strong and cooperative solutions in \emph{AdvSyftP} and \emph{CoopSyftP}, respectively. 


\noindent\textbf{Benchmarks.} For benchmarks, we consider nondeterministic planning domains as the same one in the running example of building-an-arch described in Section~\ref{sec:1-1}.
In order to have scalable benchmarks, we consider agent goals as putting $O$ objects at $L$ locations in line, hence the difficulty of the benchmarks increases as either $O$ or $L$ increases. Note that for all these benchmarks, the agent does not have a strong solution to fulfill the goal, i.e., it needs a best-effort solution. 
Our benchmarks consider at most 8 objects~($1 \leq |O| \leq 8$) and 1000 locations~($1 \leq |L| \leq 1000$). 

\noindent\textbf{Experiment Setup.}
All experiments were run on a laptop with an operating system 64-bit Ubuntu 20.04, 3.6 GHz CPU, and 12 GB of memory. Time out was set to 1200 seconds.

\begin{figure}[t]
     \centering
     \includegraphics[scale=.40]{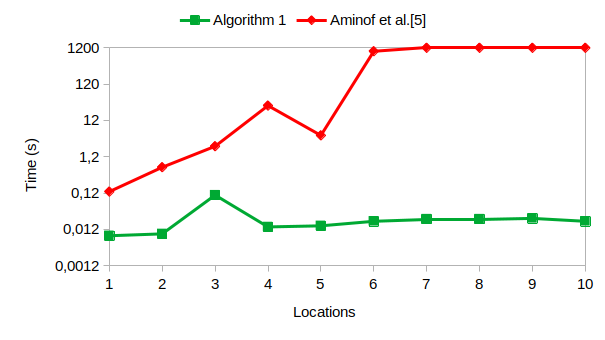}
         \caption{Comparison of \besyftp and the algorithm from Aminof et al.~\cite{ADR2021} on benchmarks with $|O| = 1$ and $1 \leq |L| \leq 10$~(in log scale).}
    \label{fig:monolithic-vs-compositional}
\end{figure}

\begin{figure}[t]
     \centering
     \includegraphics[scale=.40]{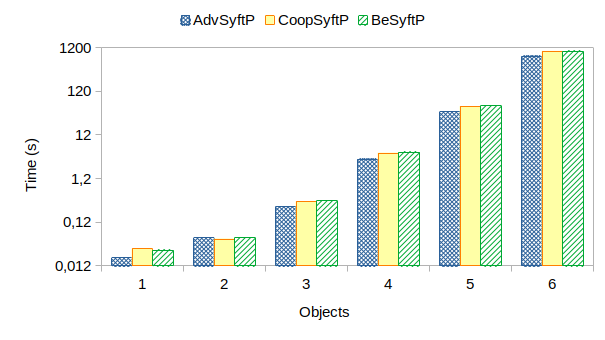}
         \caption{Comparison of \emph{BeSyftP}, \emph{AdvSyftP} and \emph{CoopSyftP} on benchmarks with $|L| = 10$ and $1 \leq |O| \leq 6$~(in log scale).}
    \label{fig:planning_comparison_1}
\end{figure}

\noindent\textbf{Experimental Results.} 
%
%
As shown in Figure~\ref{fig:monolithic-vs-compositional}~(for better vision, we only plot the results on instances with up to $10$ locations), utilizing the \LTLf best-effort synthesis approach from~\cite{ADR2021} can solve the instances with one object ($|O| = 1$) up to $6$ locations ($|L| \leq 6$).\footnote{Notice that if we increase the number of objects to $2$, the synthesis approach from~\cite{ADR2021} can still solve the corresponding instances. But if we increase the number of objects to $3$, it times out immediately.} Instead, \besyftp can solve all the instances with $|O|=1$ up to $10$ locations and more~(up to $1000$), showing much greater scalability.\footnote{If we increase $|O|$ to $6$, our approach can only handle instances with $|L|$ up to $10$. This is because $|L|$ only affects the size of the planning domain, on which the synthesis complexity is EXPTIME. Instead, $|O|$ also affects the \LTLf goal, on which the complexity is 2EXPTIME.}
%



Figure~\ref{fig:planning_comparison_1} shows the results of comparing the performance of computing best-effort solutions~(\besyftp) with that of computing strong~(\emph{AdvSyftP}) and cooperative~(\emph{CoopSyftP}) ones. Note that, since all benchmarks are \textit{unrealizable}, \emph{AdvSyftP} terminates earlier than both \emph{CoopSyftP} and \besyftp~(with $20 \% \sim 25 \%$ less time cost). Interestingly, \besyftp takes roughly the same time as \emph{CoopSyftP}~(with $10\% \sim 15\%$ more time cost). 
This shows that, although computing a best-effort solution requires solving two games and combining the computed adversarial and cooperative solutions, \besyftp requires only a small overhead with respect to both \emph{AdvSyftP} and \emph{CoopSyftP}.


\section{Conclusion}

In this paper, we have developed a framework for \LTLf best-effort synthesis in nondeterministic (adversarial) planning domains, and a solution technique that admits a scalable symbolic implementation. 
%
%
The approach presented here can be extended to different forms of goal specification. In particular, we can expect a further complexity reduction if we express goals  in pure-past temporal logics~\cite{DDFR20}. 
Also, the game-based techniques adopted here could possibly be extended to handle best-effort synthesis under multiple environment specifications~\cite{CiolekDPS20,ADLMR20,ADLMR21a}. Another promising extension is considering maximally permissive strategies~\cite{ZhuDeG22}, which allows the agent to choose a best-effort solution during execution instead of committing to a single solution beforehand 
We leave this for future work.

\section*{Acknowledgments}

This work has been partially supported by the ERC-ADG White- Mech (No. 834228), the EU ICT-48 2020 project TAILOR (No. 952215), the PRIN project RIPER (No. 20203FFYLK), and the PNRR MUR project FAIR (No. PE0000013). This work has been carried out while Gianmarco Parretti was enrolled in the Italian National Doctorate on Artificial Intelligence run by Sapienza University of Rome.

\bibliography{ecai}

\end{document}